\documentclass[iicol]{sn-jnl}

\usepackage{graphicx}
\usepackage{multirow}
\usepackage{amsmath,amssymb,amsfonts}
\usepackage{amssymb}
\usepackage{mathrsfs}
\usepackage[title]{appendix}
\usepackage{xcolor}
\usepackage{textcomp}
\usepackage{manyfoot}
\usepackage{booktabs}
\usepackage{algorithm}
\usepackage{algorithmicx}
\usepackage{algpseudocode}
\usepackage{listings}
\usepackage{adjustbox}
\usepackage{array}
\usepackage{placeins}
\usepackage{titlesec}
\usepackage{flushend}  
\usepackage{balance}   
\usepackage{hyperref}
\begin{document}

\title[Location-Free Scene Graph Generation]{Location-Free Scene Graph Generation}

\author*[1]{\fnm{Ege} \sur{Özsoy}}\email{ege.oezsoy@tum.de}
\author[1]{\fnm{Felix} \sur{Holm}}\email{felix.holm@tum.de}
\author[1]{\fnm{Mahdi} \sur{Saleh}}\email{m.saleh@tum.de}
\author[1]{\fnm{Tobias} \sur{Czempiel}}\email{tobias.czempiel@tum.de}
\author[1]{\fnm{Chantal} \sur{Pellegrini}}\email{chantal.pellegrini@tum.de}
\author[1]{\fnm{Nassir} \sur{Navab}}\email{nassir.navab@tum.de}
\author[1]{\fnm{Benjamin} \sur{Busam}}\email{b.busam@tum.de}

\affil*[1]{\orgdiv{Chair for Computer Aided Medical Procedures}, \orgname{Technical University Munich (TUM)}, \orgaddress{\city{Munich}, \country{Germany}}}

\abstract{Scene Graph Generation (SGG) is a visual understanding task, aiming to describe a scene as a graph of entities and their relationships with each other. Existing works rely on location labels in form of bounding boxes or segmentation masks, increasing annotation costs and limiting dataset expansion. Recognizing that many applications do not require location data, we break this dependency and introduce location-free scene graph generation (LF-SGG). This new task aims at predicting instances of entities, as well as their relationships, without the explicit calculation of their spatial localization. To objectively evaluate the task, the predicted and ground truth scene graphs need to be compared. We solve this NP-hard problem through an efficient branching algorithm. Additionally, we design the first LF-SGG method, Pix2SG, using autoregressive sequence modeling. We demonstrate the effectiveness of our method on three scene graph generation datasets as well as two downstream tasks, image retrieval and visual question answering, and show that our approach is competitive to existing methods while not relying on location cues. The code is available at \url{https://github.com/egeozsoy/LF-SGG}.}


\maketitle

\begin{figure*}[!t]
    \centering
    \includegraphics[width=.8\textwidth]{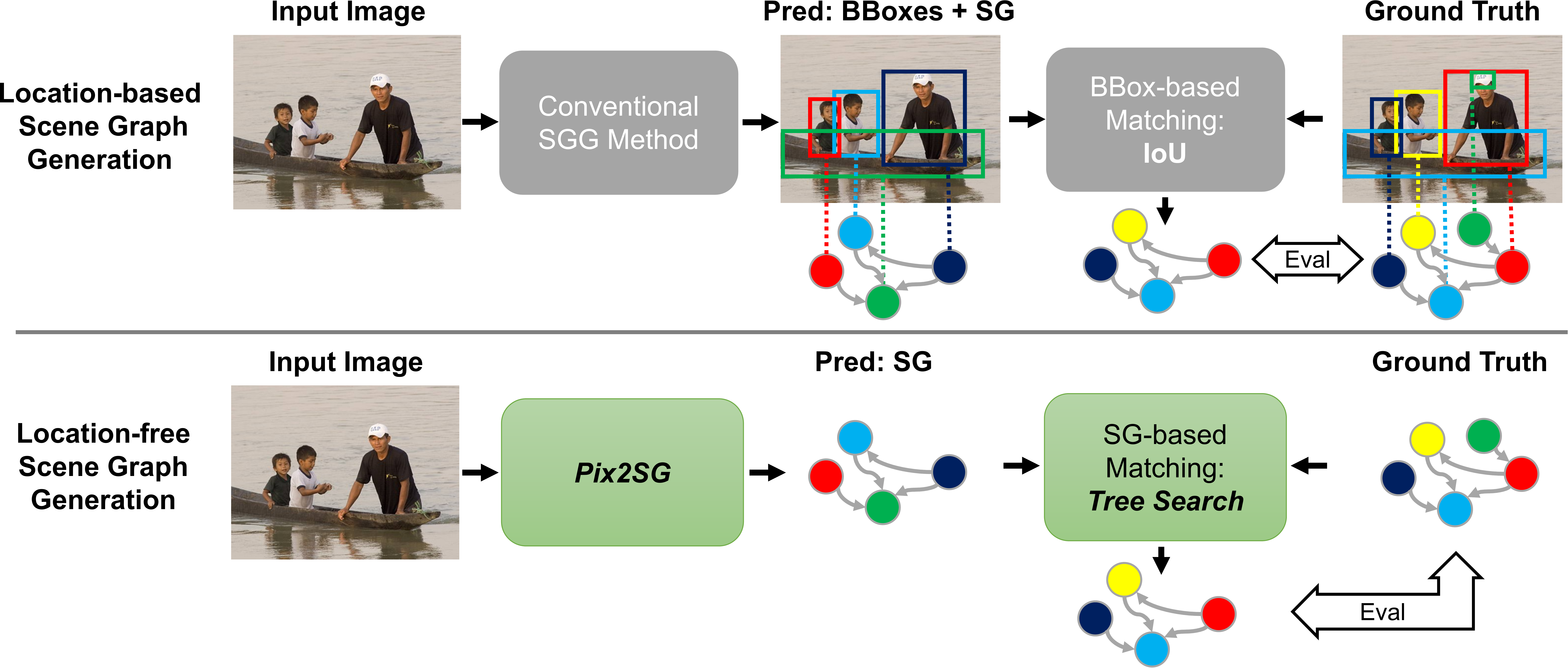}
    \caption{We introduce the new task of location-free scene graph generation (LF-SGG), completely removing the requirement for expensive bounding box or segmentation mask annotations for scene graph datasets. We further introduce Pix2SG, a method leveraging autoregressive language modeling for congruent scene graph predictions, and a heuristic tree search algorithm for scene graph matching necessary for evaluation.}
\end{figure*}

\section{Introduction}
Humans quickly abstract information from sight. The visual apparatus is our broadband interface to the world. As such, it heavily interconnects the brain to facilitate fast interpretation of sensory impressions reflected by our experience.
The task of computer-aided scene understanding reflects this process in silico with the aim of automatic visual scene interpretation from camera data. This task can be partitioned into many sub-tasks, including object and instance detection, localization, or segmentation. Much like the hierarchical abstraction in our perception, more complex tasks aim to not only describe the spatial context of objects in a scene but also try to identify the relationships between them.
Scene graphs are a potent tool for describing relationships in a scene. They provide a structured graph representation using nodes, describing the object instances and edges connecting two nodes to encode the relationship between them.
Such semantic representations of a scene can be used as the foundation for multiple downstream tasks, such as Visual Question Answering (VQA)~\cite{hudson_learning_2019}, image captioning~\cite{zhong_learning_2021}, image retrieval~\cite{johnson_image_2015}, or image generation and manipulation~\cite{dhamo_graph--3d_2021, dhamo_semantic_2020}.
While computer vision has long surpassed human performance for simpler tasks such as classification~\cite{he_delving_2015}, reliable hierarchical and efficient abstraction in the form of scene graph generation remains a challenging open research question.

Most existing scene graph generation approaches~\cite{xu_scene_2017,zhang_graphical_2019, lin_gps-net_2020, tang_unbiased_2020, zellers_neural_2018, tang_learning_2019, wu_scenegraphfusion_2021, wald_learning_2022} divide the task into two sub-tasks: 1) object detection and 2) pairwise relationship prediction. Recent works also propose to use an end-to-end pipeline, where object detection and relationship prediction are performed and optimized jointly~\cite{teng_structured_2022, cong_reltr_2022, shit_relationformer_2022}. 

Nonetheless, to our knowledge, all previous methods require object locations, such as bounding boxes or segmentation masks, in some part of their pipeline. However, object locations are not needed for many downstream applications such as image retrieval, visual question answering, or image captioning~\cite{hudson_learning_2019, zhong_learning_2021, johnson_image_2015, dhamo_graph--3d_2021, dhamo_semantic_2020, yang2019auto, gu2019unpaired} but are mainly used to simplify the SGG pipeline. Their location information provides a straightforward way for instance-level matching in case multiple objects of the same type are present in the scene. However, bounding boxes and semantic segmentation labels add significant overhead in creating datasets and can further include annotation errors.

In our work, we challenge the requirement of object localizations for scene graph generation and evaluation and overcome the burden of location labels. Freeing the task from location annotations could not only significantly simplify the creation of scene graph datasets but it would also allow to disentangle the research fields of object detection and scene graph generation.
While there are many benefits of predicting scene graphs without object location annotations, existing approaches face two major pitfalls. Firstly, the detection task is deeply rooted in the design of most methods, and its removal makes the scene graph generation task impossible for these pipelines. Secondly, current scene graph evaluation metrics are unable to evaluate the performance without bounding box or mask information. The metrics measure the intersection over union between prediction and ground truth based on their locations and use this in the subsequent evaluation of the predicates~\cite{xu_scene_2017}. Without localization, it is not possible to evaluate the predicates for multiple different instances.

In this paper, we introduce the new task of location-free scene graph generation (LF-SGG) and make its objective evaluation feasible. We design the first baseline method, which overcomes the challenge of correctly identifying entities with their corresponding instances using a novel transformer-based sequence generation design, which does not rely on location annotations during training or validation. Our method can represent the complexities in an image or scene and produce a corresponding scene graph. This scene graph provides a structured and parsable abstraction of the scene, even without precise pixel-wise location information. Even though we neither train nor evaluate our model on the object location, our method intrinsically learns location awareness, which could be retrieved using the transformer's attention maps. By design, our approach abstracts spatial information into relational and semantic descriptors such as "close", "in front", or "over", maintaining the essential spatial context necessary for understanding object relationships, without relying on precise localization details. We show the usefulness of location free scene graphs in two downstream tasks, namely image retrieval and visual question answering (VQA).

We further introduce the first location-free evaluation paradigm for SGG. The predicted scene graph is matched with the target graph using a heuristic tree search formulation. This provides a task-adjusted approximate solution to the NP-hard~\cite{gold_graduated_1996} matching problem of two scene graphs and thereby enables objective evaluation of the new task.

Extensive experiments on the PSG~\cite{yang_panoptic_2022}, Visual Genome~\cite{krishna_visual_2016} and 4D-OR~\cite{ozsoy_4d-or_2022} datasets underscore the performance of the proposed model on this new task and validate our design choices. On PSG, our method outperforms every method except one, and on Visual Genome, our model performs only slightly worse than the current state-of-the-art methods that rely on significantly more labels during training~\cite{xu_scene_2017, tang_learning_2019, teng_structured_2022, tang_unbiased_2020, zellers_neural_2018}. On 4D-OR~\cite{ozsoy_4d-or_2022}, we even surpass the location-based state-of-the-art method. Further, our location free scene graphs beat the baseline methods on image retrieval, and perform on par with traditional scene graphs on VQA.

\noindent In summary, we make the following major contributions:

\begin{itemize}
\item We propose the new task of location-free scene graph generation (LF-SGG). The challenge is to predict scene graphs without access to any location information at training or evaluation time.
\item A necessity for the objective evaluation of LF-SGG is scene graph matching. To this end, we design a task-specific efficient heuristic tree search algorithm.
\item We introduce Pix2SG, the first method to solve LF-SGG.
\item We perform extensive experiments on three scene graph generation datasets, PSG, Visual Genome and 4D-OR, and on two downstream tasks (image retrieval and VQA) to validate both location free scene graphs as well as our method.
 
\end{itemize}

\section{Related Work}
\subsection{SGG with Location Supervision}
Previous methods for SGG have typically relied on two-stage architectures. The first stage consists of an object detector, often a pretrained Faster-R-CNN model~\cite{ren_faster_2016}.
The detected localized objects are used as proposals for the scene graph. The proposed objects and visual relationships between them are then classified in the second stage of the architecture, which can take the form of Iterative Message Passing~\cite{xu_scene_2017, zareian_weakly_2020}, Graph Neural Networks~\cite{zhang_graphical_2019, lin_gps-net_2020, yang_graph_2018, wald_learning_2020, wu_scenegraphfusion_2021}, and Recurrent Neural Networks~\cite{zellers_neural_2018, tang_learning_2019}.
Lately, some works have tried to move away from this paradigm towards end-to-end approaches~\cite{teng_structured_2022, cong_reltr_2022, shit_relationformer_2022, yang_panoptic_2022}, closely integrating the generation of localized objects and their relationships. These methods are sometimes described as detection-free because they omit an explicit object detector, but they still train to localize the entities in the output scene graph. All of the methods mentioned above rely on location supervision in the form of bounding box or mask labels, regardless of architectural choice.

\subsection{Weakly-Supervised Scene Graph Generation}
Recently, some works have attempted Weakly-Supervised Scene Graph Generation (WS-SGG), omitting the costly location labels in their training. Zareian et al. (VSPNet~\cite{zareian_weakly_2020}) take object proposals from a pretrained object detector, build them into a semantic bipartite graph as a different formulation of a scene graph and classify and refine the entities through message passing. Shi et al.~\cite{shi_simple_2021} use a weakly-supervised graph matching method to match object proposals from a pretrained object detector to the ungrounded scene graph labels. This matching generates a localized scene graph as a pseudo-label to train conventional fully-supervised SGG methods. Other methods utilize natural language from image captions as a weak supervision source, matching linguistic structures to object proposals that again come from pretrained object detectors~\cite{zhong_learning_2021, ye_linguistic_2021, li_integrating_2022}.

We argue that the reliance of the WS-SGG methods mentioned above on pretrained object detectors is still a major limitation. While this approach does not require location labels in the scene graph datasets itself, a dataset from the same domain that contains location labels for all relevant objects is still required to pretrain the object detector. Shi et al.~\cite{shi_simple_2021} notice a considerable domain gap for the detector pre-training. This manifests in subpar performance if Faster-R-CNN is pretrained on Open Images~\cite{kuznetsova_open_2020} and applied to Visual Genome~\cite{krishna_visual_2016} data, although both contain natural images. This issue becomes even more apparent when SGG is attempted in domains where large-scale image datasets are not readily available, such as the medical domain~\cite{ozsoy_4d-or_2022}. We instead enable scene graph generation without the use of pretrained object detectors to eliminate the need for location supervision completely.

The methods mentioned in this section output localized scene graphs without training on location labels and are therefore considered weakly supervised. While our suggested LF-SGG task also does not include location labels, it further works without an intermediate generation of localized scene graphs. We, therefore, do not consider LF-SGG weakly supervised.

\subsection{Scene Graph Matching}

Fully supervised location-based scene graph generation methods rely on the location to match the scene graph nodes of the prediction and ground truth graph for evaluation by simply thresholding bounding box or mask IoU~\cite{xu_scene_2017}. When location-free (ungrounded) scene graphs are used, this simple solution for graph matching is no longer available.
The WS-SGG methods shown in the previous section also address some graph-matching problems. VSPNet~\cite{zareian_weakly_2020} aligns their predicted semantic bipartite graph to the ground truth with an iterative optimization algorithm. Shi et al.~\cite{shi_simple_2021} leverage the simple and efficient Hungarian matching algorithm~\cite{kuhn_hungarian_1955} to find a first-order graph matching of node to node without considering graph structure. Both of these approaches, however, rely on visual features extracted from the used object detector to find a strong node-to-node similarity. In location-free scene graphs visual features are no longer available. We, therefore, introduce a new scene graph matching algorithm that is capable of matching scene graphs purely based on node labels, edge labels, and graph structure by heuristic tree search.

\subsection{Autoregressive Decoding}
Previous SGG methods generally leverage object proposals to initialize a scene graph. Direct prediction of ungrounded scene graphs at once from an image without this intermediary step requires an highly complex output vector. This could impede training and lack the potential to model the interdependencies between the entities and relations of the scene graph. 

We, therefore, look at autoregressive decoding, which allows a model to make multiple predictions from one data object sequentially. Given the similarity of scene graphs to texts regarding grammatical and semantic structure, we take inspiration from natural language processing (NLP).
Transformers and autoregressive decoding have enabled great progress in this domain lately~\cite{devlin_bert_2019, alayrac_flamingo_2022, brown_language_2020}. Chen et al. (Pix2Seq~\cite{chen_pix2seq_2022}) showed that autoregressive decoding could also be used successfully with images for object detection. We argue that we can leverage the advantages of autoregressive decoding even more so in the field of SGG, where there are significant interdependencies in the semantic structure of the scene graph, which can profit from more congruent sequential predictions. We introduce Pix2SG, a model featuring autoregressive sequence decoding for LF-SGG.

\section{Method}

In this section, we first introduce the new location-free scene graph generation task. To our knowledge, all existing SGG methods require location information, in the form of bounding boxes or segmentation masks in some parts of their pipeline, making them unsuitable for the LF-SGG task. As to our knowledge all prior work relies on location information, in Sec.~\ref{sec:prop_solution}, we present our novel architecture, specifically designed for LF-SGG. Lastly, we design a heuristic tree search-based matching algorithm to enable objective evaluation of our new task.

\subsection{Problem Formulation}
\noindent
In this section, we define location-free scene graph generation as the prediction of a scene graph given an image of a scene without using location. The input is an image $I$, and the goal is to produce a scene graph prediction $G = (V,E)$, where $V$ corresponds to the entity nodes, and $E$ to the pairwise relationships. For an objective evaluation, the matching
\begin{equation} \label{eq:graph_matching}
M\left( G,G' \right) = G_{m}
\end{equation}
 fits the prediction $G$ to the ground truth $G'$, to produce the mapped graph prediction $G_{m}$.
Recall@K $\mathcal{R}_K \left( G_{m},G' \right)$ is computed between $G_{m}$ and $G'$ to evaluate the task. With this formulation, unlike the conventional SGG task, the LF-SGG task refrains from using location information for both subject and object entities. The task of LF-SGG is more challenging compared to SGG, where the location labels provide a strong supervision signal.

\subsection{Proposed Solution}
\label{sec:prop_solution}

\begin{figure}[!t]
  \centering
  
   \includegraphics[width=0.95\linewidth]{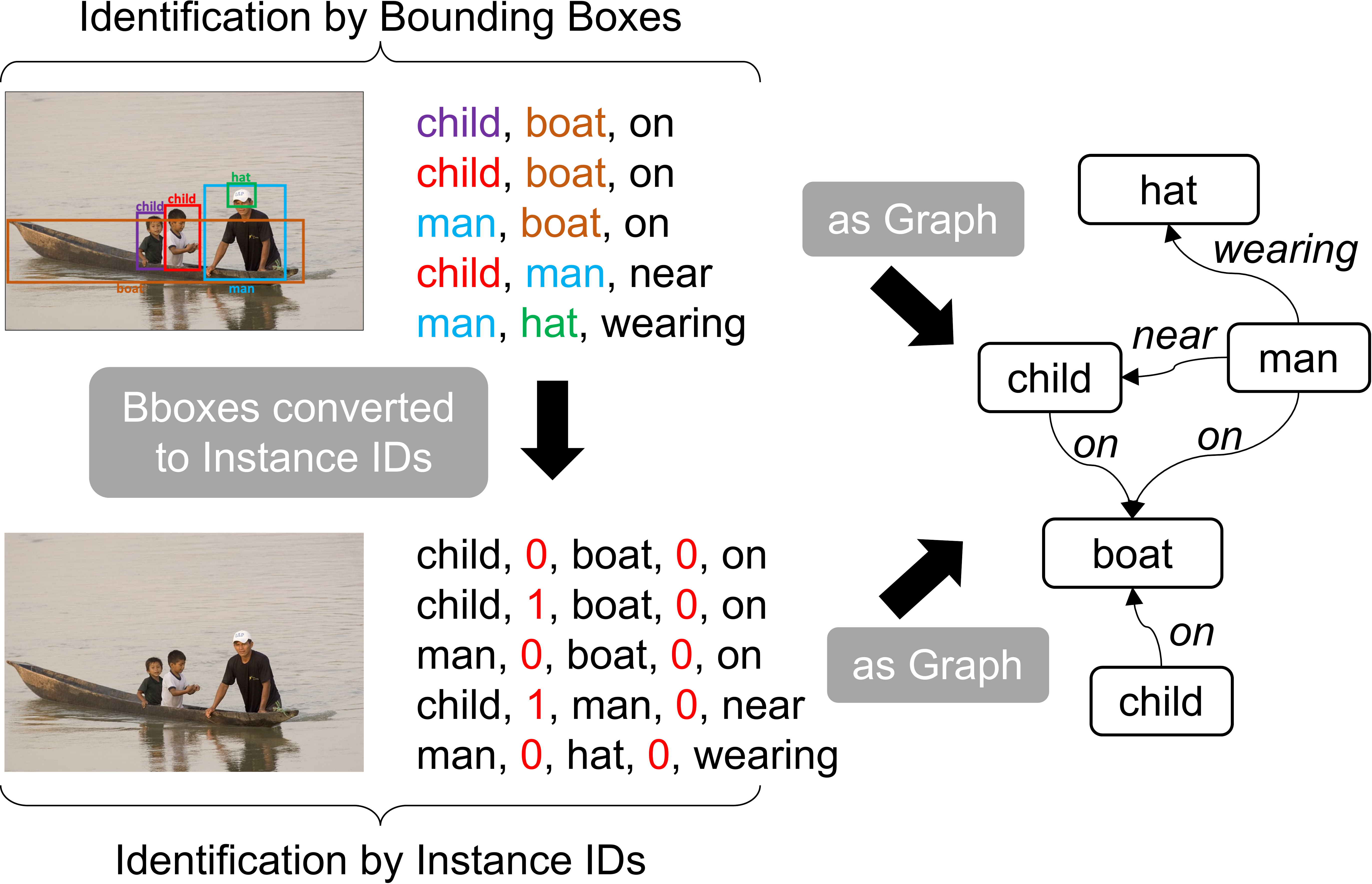}

   \caption{Conversion of existing location-based scene graph annotations to location-free scene graphs with instance identification and mapping to the graph representation.}
   \label{fig:sg_no_loc}
\end{figure}

\begin{figure*}[!t]
  \centering
  
   \includegraphics[width=0.95\linewidth]{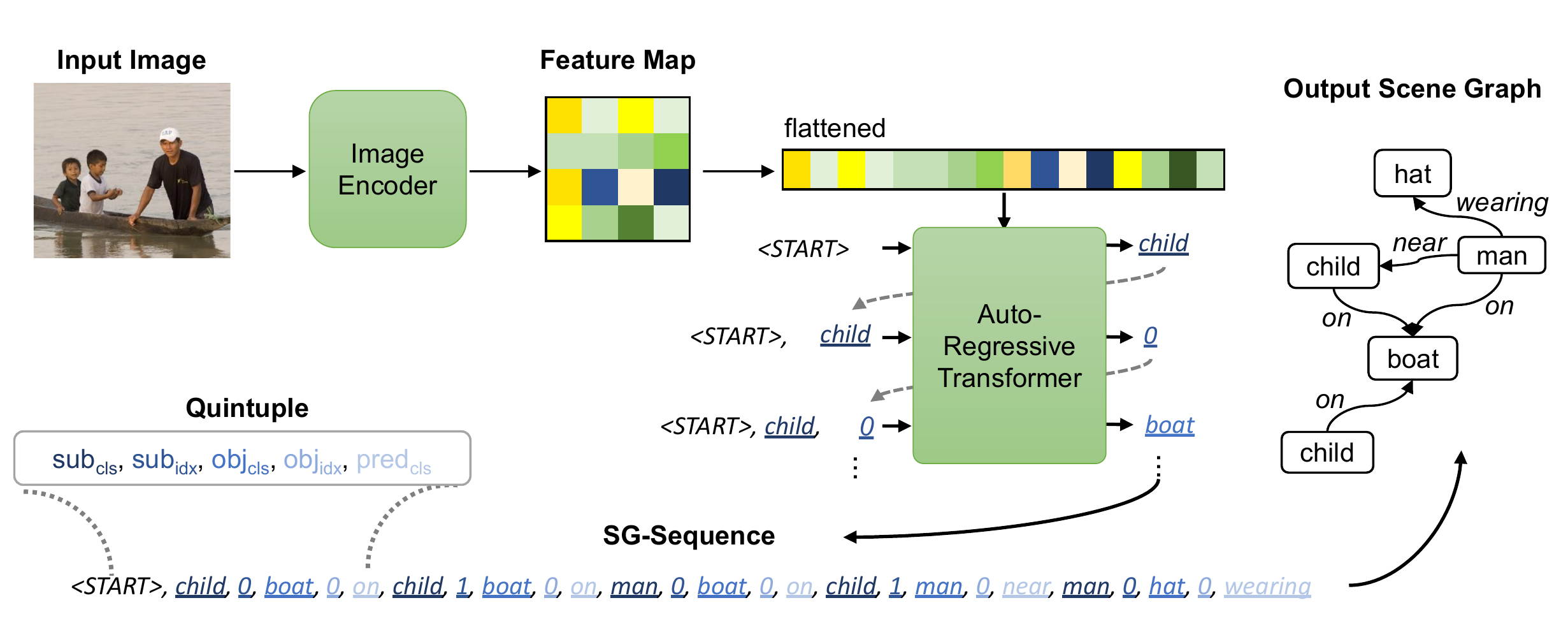}

   \caption{Pix2SG Architecture: An image encoder encodes the image as a feature map that is flattened and used as the input sequence to the autoregressive transformer module. The autoregressive transformer predicts the components of the scene graph, token by token, considering all its previous predictions until the output SG-sequence is completed.}
   \label{fig:autoreg_sg}
\end{figure*}

As a first solution to LF-SGG, we introduce our proposed architecture, Pix2SG, the first location-free scene graph generation method.
Inspired by Pix2Seq~\cite{chen_pix2seq_2022}, Pix2SG follows the autoregressive pattern, which is a well-known paradigm in natural language processing, but so far has not been used for the task of scene graph generation. We simultaneously predict objects as entity labels and entity instance indices, as well as pairwise relationships directly from a given image. Our model produces the entire output in a straight-forward, autoregressive fashion, predicting one token after another until the entire scene is described, as visualized in Fig.~\ref{fig:autoreg_sg}. One advantage of this autoregressive formulation is that it allows our network to exploit the dependencies and relations between different tokens and triplets to improve its prediction. 

\noindent\textbf{Vocabulary.} To formulate scene graph generation as autoregressive sequence generation, we first define a vocabulary of tokens uniquely identifying all entities and predicates in the scene graph dataset. Each entity is represented by two tokens, where the first token represents the entity class and the second token the entity instance, differentiating entities of the same class in one image. Additionally, we represent each predicate with a single token $pred_{cls}$. Therefore, a relation between two entities can be encoded using our vocabulary as a quintuple with a combination of five tokens, 
\begin{equation} \label{eq:SceneGraphQuintupel} (sub_{cls},sub_{idx},obj_{cls},obj_{idx},pred_{cls}) \end{equation}
where $sub_{cls}$ and $obj_{cls}$ represent the entity class and $sub_{idx}$ and $obj_{idx}$ the instance ids of the subject and object. 

\noindent\textbf{Ground Truth Sequence Generation.} 
To train our proposed sequence generation model, we convert a scene graph label $G'$ into a sequence of tokens from our vocabulary. This SG-sequence will be used in both the training and inference of our method.
We first convert the scene graph into individual quintuples as visualized in Fig.~\ref{fig:sg_no_loc}, and then randomly concatenate all quintuples to generate the SG-sequence. We assign ascending entity IDs to multiple instances of the same class, where the first appearance of a class in the sequence will have the instance id 0, the next one 1, and so on. If the same instance occurs in another quintuple, it is assigned the same instance id. We design our evaluation method to be invariant to the order of the quintuples as well as to the order of the instance ids to eliminate any ambiguities during validation.

\noindent\textbf{Network.} Our approach, visualized in Fig.~\ref{fig:autoreg_sg}, first employs an image encoding step to generate a flattened representation of the image features encoding the visual information of the frame. In accordance with previous works~\cite{chen_pix2seq_2022}, we augment the flattened image information with a positional encodings which allows the model to identify the spatial location of each token in the feature map. The flexibility of our approach allows us to use any feature extraction backbone, and does not restrict us to approaches trained for object detection or segmentation.

In the decoding step of the method, the flattened image feature and the start token are used as the first input to the decoder network to generate the first output token. Each predicted output token is subsequently appended to the token sequence that accumulates the predicted tokens. This token sequence is then again used as input to generate the next output token in an autoregressive manner. By the repetition of this autoregressive step, our method is generating an output sequence. This sequence of tokens can then be effortlessly translated into the desired scene graph representation (Fig.~\ref{fig:sg_no_loc}).

\noindent\textbf{Inference.} 
During inference, the next predicted token of the SG-sequence should intuitively be selected using the confidence values of the prediction. However, we noticed that this can lead to unwanted repetitions of the same token in the SG-sequence. To address this problem, we propose the use of nucleus sampling~\cite{holtzman_curious_2020}, which introduces a limited amount of randomness for the selection of the next token. For the experiments, we always generate a fixed number of tokens and convert the results into an output of $N$ SG-quintuples (Eq.~\ref{eq:SceneGraphQuintupel}) with $N$ defining the number of total relations. Our autoregressive approach already encourages predicting the most prominent relationships first, before predicting more specific ones later in the sequence. Furthermore, if desired, our model is capable of predicting a "stop token", to naturally stop the scene graph generation when an appropriate level of coverage of the scene is reached. This level ultimately is learned from training data distribution, similar to location-based scene graph generation as well as image captioning.

\subsection{Objective Evaluation Process}

\begin{figure*}[!t]
  \centering
   \includegraphics[width=1\linewidth]{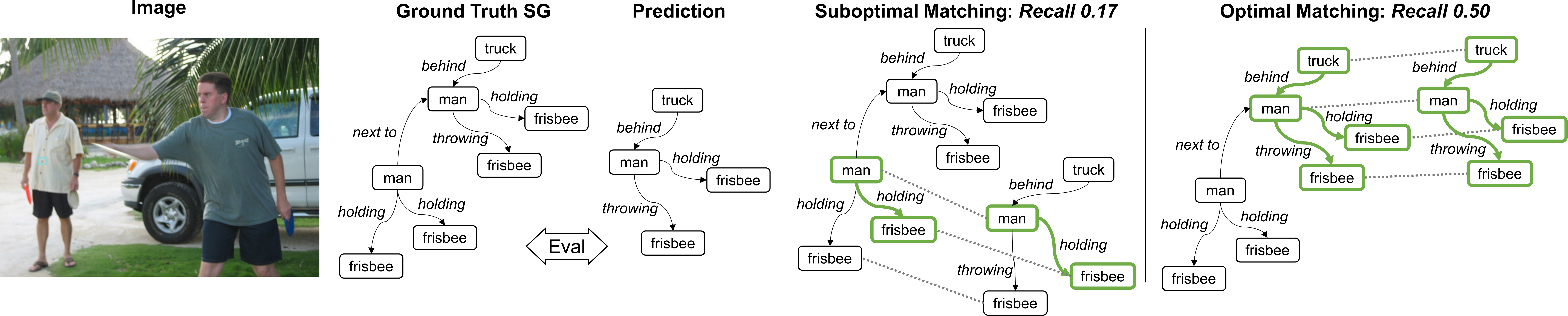}
   \caption{Illustration of the scene graph matching problem. Ground truth scene graph and prediction have to be correctly matched for the evaluation. A suboptimal matching can obscure the model actual performance.}
   \label{fig:sg_matching_problem}
\end{figure*}

To facilitate an objective evaluation process for LF-SGG, scene graph matching has to be performed to calculate $M\left( G,G' \right) = G_{m}$ (Eq.~\ref{eq:graph_matching}). The evaluation of LF-SGG has to be performed by only relying on the graph information. However, as shown in Fig.~\ref{fig:sg_matching_problem}, matching graphs with each other is not straightforward, especially when multiple class instances are present in one scene. In this case, node identification has multiple possible solutions, and identifying the correct instance can only be done by considering the entire graph. Finding the optimal correspondence between prediction and ground truth involves calculating all possible combinations for the matching, maximizing the overlap of graph nodes and relationships. On the one hand, an exhaustive search over all possible matching combinations is computationally very expensive. On the other hand, the Hungarian matching algorithm~\cite{kuhn_hungarian_1955} cannot address the task appropriately, as the cost of an assignment can not be directly derived from the vicinity of each node but only after a complete matching. 

Therefore, we propose a new algorithm shown in Alg.~\ref{alg:tree_based_matching} to approximate an exhaustive search with an emphasis on 1) matching quality, 2) computational efficiency, and 3) flexibility and configurability. Our matching algorithm $M^{\ast}\left( G,G',B \right) = G_{m}^{B}$ retrieves an approximate solution $G_{m}^{B}$ to the matching problem $M$ (Eq.\ref{eq:graph_matching}). It is inspired by tree search algorithms. We define a branching factor $B$ to control the depth of our algorithmic tree search. By setting $B$ to 1, our algorithm performs greedy matching where each predicted entity instance is iteratively mapped to the ground truth entity instance with the highest overlap. To compute the overlap, we first get the respective local one-hop neighborhoods, represented as a list of (predicate, entity class) tuples. We count how many of these tuples are the same for both of them, where both the predicate and the entity classes have to match. This score is normalized by the node degree. For $B > 1$, the $B$ highest overlaps will be used to create new branches, representing multiple alternative mappings. These branches will be explored recursively until all the instances are mapped. For $B \geq N$, where $N$ is the number of instances of the most common entity class in a scene, our algorithm performs an exhaustive search to find the exact but computationally expensive solution. It holds $G_{m}^{B} \xrightarrow{B \to N} G_{m}$. To further optimize the performance, we start our tree search from the ground truth entity node with the highest node degree, arguing that this node will likely be a key component of our scene.

\begin{algorithm}
\caption{Heuristic Tree Search (HTS)} 
\textbf{Input:} gt graph $y$, predicted graph $\hat{y}$, branching factor $B$\\
\textbf{Output:} best mapping $m_{best}$ \\
\textbf{function} HTS($y, \hat{y}, B, m$)
\begin{algorithmic}
\State $y\_inst \gets$ instance from $y$ with highest node degree
\State $\hat{y}\_insts \gets $ instances from $\hat{y}$ with same class as $y\_inst$
\State $y\_nbhd \gets$ connected nodes and edges of $y\_inst$
\For{$\hat{y}\_inst$ in $\hat{y}\_insts$}
\State $\hat{y}\_nbhd \gets$ connected nodes and edges of $\hat{y}\_inst$
\State $overlaps \gets $ append $|y\_nbhd \cap \hat{y}\_nbhd| / |y\_nbhd|$
        \EndFor
    \State $\hat{y}\_insts \gets \hat{y}\_insts$ sorted by $overlaps$
        \State $M \gets \varnothing $  set for branched mappings $m_i$
    \For{$i = 0$ : $B$}{~$m_i \gets m \cup (y\_inst\mapsto$ $\hat{y}\_insts[i]) $}
        \If{$y\backslash y\_inst == \varnothing$} $M \gets M \cup m_i$
        \Else{~$M \gets M \cup $ $\small\text{HTS}(y\backslash y\_inst, \hat{y}\backslash \hat{y}\_insts[i], B, m_i)$}    
        \EndIf
   \EndFor\\
\textbf{return} $M$

\end{algorithmic}
$m_{best} \gets$ select highest recall from $\small\text{HTS}(y, \hat{y}, B, \varnothing$)
\label{alg:tree_based_matching}
\end{algorithm}

The result of our heuristic tree search is a set of graph matches between prediction and ground truth, and we subsequently select the match producing the highest evaluation metric as visualized in Fig.~\ref{fig:sg_matching_problem}. We then use this matching to convert $G$ to $G_{m}$.

\section{Experiments}
\subsection{Datasets}
\noindent\textbf{Visual Genome (VG).} The most frequently used scene graph dataset is VG~\cite{krishna_visual_2016} and it is considered as one of the main benchmarking datasets for SGG. As most previous works~\cite{teng_structured_2022}, we use a split of VG with the 150 most frequent objects and 50 predicates. While conventionally PredCls, SGCls, and SGGen are used as metrics, none are applicable to the case of LF-SGG. Instead, we evaluate all approaches with our proposed metric, LF-SGGen, where the recall is calculated directly by matching and comparing the predicted scene graph to the ground truth scene graph.

\noindent\textbf{Panoptic Scene Graph Dataset (PSG).} PSG~\cite{yang_panoptic_2022} is a more recent dataset, designed for panoptic scene graph generation, where location information is not provided as bounding boxes, but as more accurate segmentation masks. Merging aspects of COCO and Visual Genome (VG), PSG consists of 49k images, annotating 133 objects and 56 unique predicates. They address some shortcomings of Visual Genome, such as redundant class and predicate labels, annotation of trivial relationships as well as duplicate localizations. The improvements over Visual Genome make it a more interesting benchmark for evaluating scene graph generation. 

\noindent\textbf{4D-OR.} 4D-OR~\cite{ozsoy_4d-or_2022} is a recently published surgical scene graph dataset. Unlike VG and PSG, which are sparse in their annotations, 4D-OR includes dense annotations, enabling the calculation of precision in addition to recall. As it includes images from multiple views per scene, it allows us to demonstrate the extension of our method for multiple image inputs per scene. Finally, as the dataset size is an order of magnitude smaller than VG and PSG, it allows us to evaluate the performance of our method in lower data regimes. 

\subsection{Downstream Tasks}

\noindent\textbf{Image Retrieval.} We evaluate the usefulness of our location free scene graphs in the task of image retrieval, specifically using the Sentence-to-Graph Retrieval (S2GR) methodology introduced by \cite{tang_unbiased_2020}. S2GR first converts image captions into scene graphs, and learns to match them with image scene graphs. The goal is to find the correct image, given an image pool of 1000 or 5000 images, measured using R@20 and R@100. This task is deliberately designed to avoid using image features, and instead only focuses on the graphs. We follow their implementation very closely, and only replace their image scene graphs by our own location free scene graphs generated using Pix2SG VIT-L. 

\noindent\textbf{VQA.} We further evaluate our LF-SGs for the task of visual question answering. Concretely, we use the COCOVQA~\cite{visual_question_answering} dataset, which consists of multiple question answer pairs for each image. While existing methods rely directly on image features and supervised training, our task is zero-shot, and does not use any labeled data for VQA. We start by using a SGG method to compute scene graphs for the images, and then feed the scene graphs as text (list of triplets) into a Large Language Model~(LLM) as a part of the prompt for question answering. We use the following prompt: "Given is the following scene graph for an image: \textless SG\textgreater. Answer the following benchmark question. If you are unsure, make an educated guess. Don't give explanations, your output will be automatically evaluated. You try to maximize your score on the benchmark. Answers mostly consists of one or two words, be concise. Question: \textless Q\textgreater", where we replace \textless SG\textgreater~and \textless Q\textgreater~with the respective image scene graph and question. This way, an existing LLM can be directly utilized for reasoning about an image. We evaluate the accuracy of the SGG methods on all categories, and the three subcategories, open end, number, and yes/no.  

\subsection{Implementation details}
\begin{figure}[!t]
  \centering

  \includegraphics[width=0.9\linewidth]{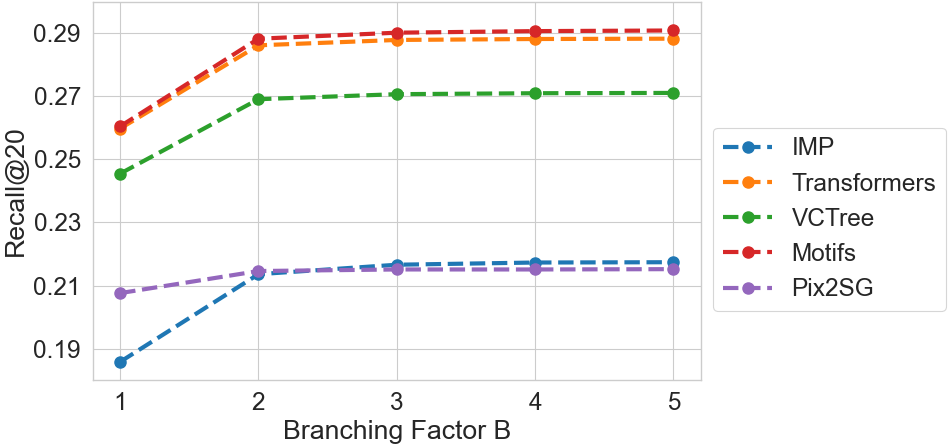}

  \caption{Ablation of Branching factor with $B=3$ providing a good trade-off between speed and matching performance.}
  \label{fig:convergence_20}
\end{figure}

We use both EfficientNet~\cite{tan_efficientnet_2020} pretrained on Imagenet~\cite{russakovsky_imagenet_2015} as well as Vision Transformer~\cite{dosovitskiy2020image} with contrastive pretraining~\cite{radford2021learning} as  image encoder backbones. We resize the images to match the input dimensions of the backbones. In 4D-OR, the four multi-view images per scene are processed individually by the feature extraction backbone, then the feature maps are flattened and concatenated to build the input sequence. We use pix2seq~\cite{chen_pix2seq_2022} as the starting point of the autoregressive sequence modeling implementation.
We use a categorical cross-entropy loss, with the entire vocabulary as target classes, and optimize our model with AdamW~\cite{loshchilov_decoupled_2019} and a constant learning rate of $4 \times 10^{-5}$ with weight decay of $1 \times 10^{-4}$. The batch size is set to 16 in all our experiments and we train our methods for 200 epochs, employing early stopping. We use a Transformer~\cite{vaswani_attention_2017}, with a hidden size of 256, eight attention heads, six encoding, and two decoding layers. Unless otherwise specified, we predict 300 relations using nucleus sampling~\cite{holtzman_curious_2020} with a p-value of 0.95 and pick the top K unique predictions for Recall@K. We set the branching factor $B$ of our proposed heuristic three-matching algorithm to 3 for all the validation experiments except when indicated otherwise. For the baseline methods on PSG, Visual Genome, and 4D-OR, we use the implementations provided by~\cite{yang_panoptic_2022}, ~\cite{tang_scene_2020} and ~\cite{ozsoy_4d-or_2022} respectively. Finally, we provide an efficient implementation of our Heuristic Tree Search based evaluation algorithm in C++, with a sub-second run time for most samples using three as Branching-factor $B$. We empirically motivate the choice of $B$ in Sec.~\ref{sec:ablationstudies} and Fig.~\ref{fig:convergence_20}. All of our experiments are done on a single NVIDIA A40 GPU.

\subsection{Results}
\label{sec:results}
\begin{figure*}[t]
  \centering

  \includegraphics[width=0.95\linewidth]{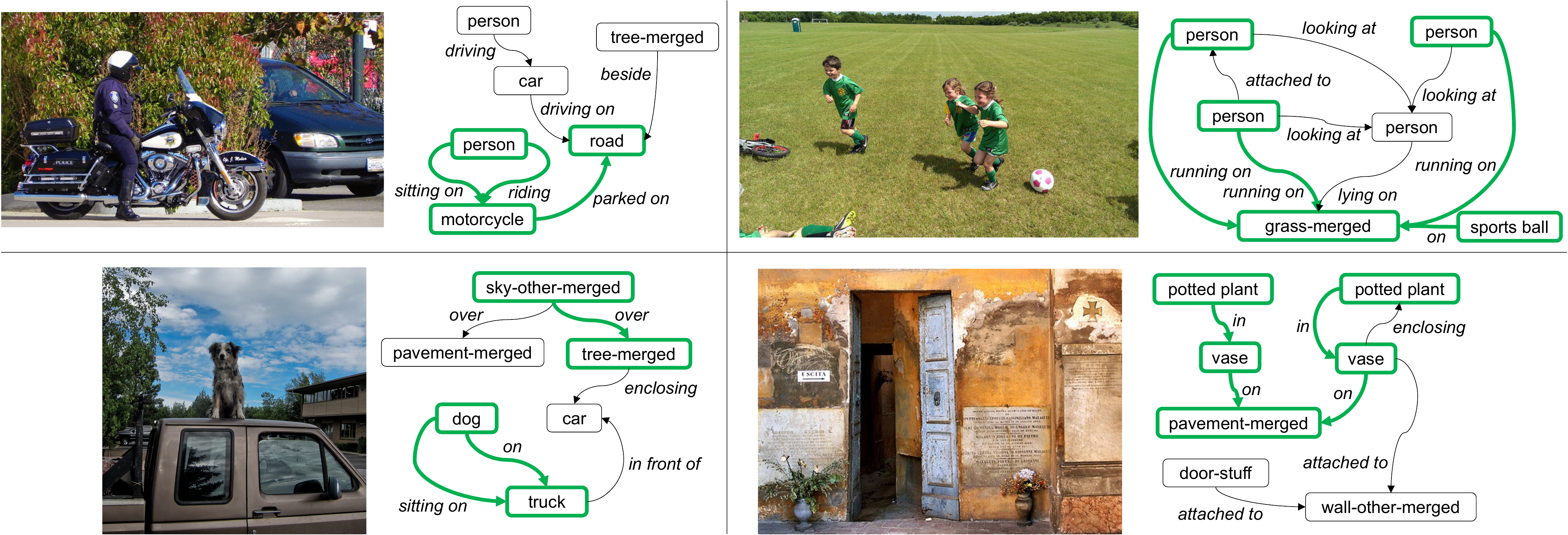}
  \caption{Qualitative Results of Pix2SG on the Panoptic Scene Graph Dataset. Images and corresponding Ground Truth Scene Graphs are shown. Nodes and edges correctly predicted by our model are highlighted in green. Additional triplets are predicted which are not in the ground truth but meaningful.}
  \label{fig:qual_psg}
\end{figure*}
\begin{figure*}[t]
  \centering

  \includegraphics[width=0.9\linewidth]{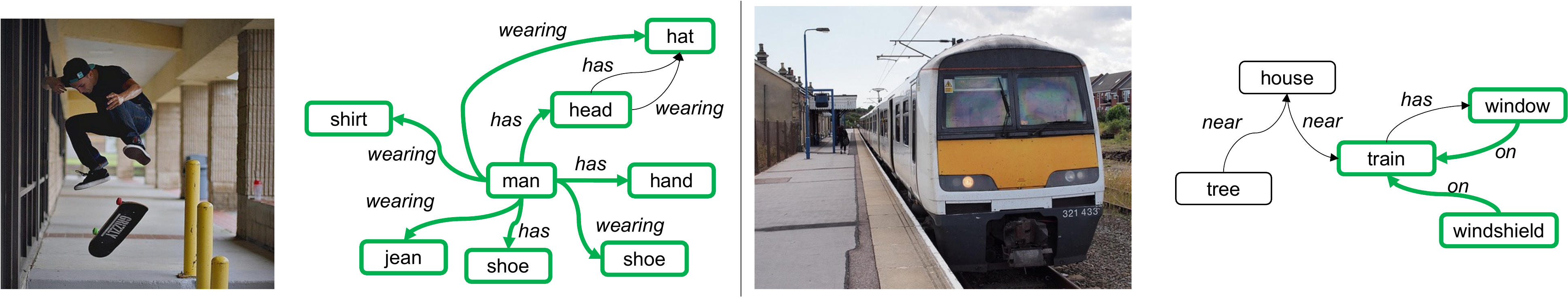}
  \caption{Qualitative Results of Pix2SG on the Visual Genome dataset. Images and corresponding Ground Truth Scene Graphs are shown. Nodes and edges correctly predicted by our model are highlighted in green. Many correctly classified predicates are of geometrical nature even though our method does not include the localization task.}
  \label{fig:qual_vg}
\end{figure*}

\begin{figure*}[!t]
  \centering
  \includegraphics[width=0.8\linewidth]{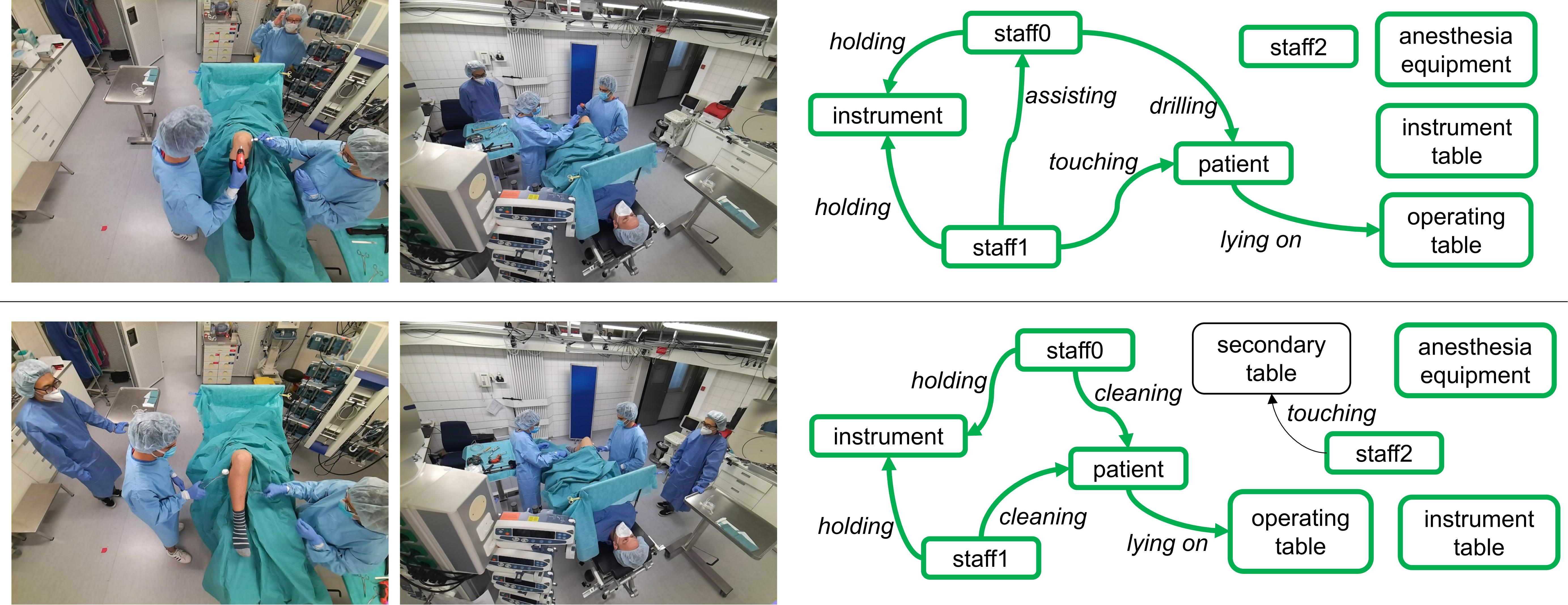}
         \caption{Qualitative Result of Pix2SG on 4D-OR dataset. Images from two of the six viewing angles and the corresponding ground truth scene graphs are displayed. Nodes and edges correctly predicted by our model are highlighted in green.}
  \label{fig:qual_4dor}
\end{figure*}

\begin{table*}
  \centering
    \caption{LF-SGGen results of different SG models at R@k on PSG dataset. B5, B7, VIT-B, VIT-L represent the corresponding EfficientNet and Vision Transformer backbones we used in our model.}
  \label{tab:psg_results}
  \begin{tabular}{ l |  c | c c c}
  \toprule
    Model & Location-Free & R@20 & R@50 & R@100 \\
    \midrule
    IMP~\cite{xu_scene_2017}  &  & 25.38 & 29.46 & 31.06 \\
    GPSNet~\cite{lin_gps-net_2020} &  & 25.96 & 30.03 & 31.91 \\
    PSGFormer~\cite{yang_panoptic_2022} & & 26.10 & 31.47 & 34.75 \\
    MOTIFS~\cite{zellers_neural_2018} & & 30.50 & 34.68 & 36.30 \\
    VCTree~\cite{tang_learning_2019} & & 31.72 & 36.28 & 37.99 \\
    PSGTR~\cite{yang_panoptic_2022} & & \textbf{39.25} & \textbf{43.95} & \textbf{44.11} \\
    \midrule
    Pix2SG B5(Ours) & $\checkmark$ & 29.07 & 34.63 & 37.00\\
    Pix2SG B7(Ours) & $\checkmark$ & 30.66 & 34.28 & 35.92\\
    Pix2SG VIT-B(Ours) & $\checkmark$ & 33.35 & 38.10 & 39.93\\
    Pix2SG VIT-L(Ours) & $\checkmark$ & \textbf{35.54} & \textbf{40.40} & \textbf{41.72}\\
    \bottomrule
  \end{tabular}  
\end{table*}

\begin{table*}
  \centering
    \caption{LF-SGG results of different SG models at R@k on Visual Genome dataset. B5, B7, VIT-B, VIT-L represent the corresponding EfficientNet and Vision Transformer backbones we used in our model.}
  \label{tab:vg_results}
  \begin{tabular}{ l | c | c c c}
  \toprule
    Model & Location-Free  & R@20 & R@50 & R@100 \\
    \midrule
    IMP~\cite{xu_scene_2017} &  & 21.66 & 30.78 & 37.07 \\
    SS-R-CNN~\cite{teng_structured_2022} & & 22.09 & 26.43 & 28.57 \\
    SGTR~\cite{li_sgtr_2022} &  & 23.62 & 30.38 & 34.85 \\
    RelTR~\cite{cong_reltr_2022} &  & 25.86 & 30.99 & 33.31 \\
    VCTree~\cite{tang_learning_2019} &  & 27.06 & 35.59 & 41.21 \\
    Transformer~\cite{tang_unbiased_2020} &  & 28.79 & 37.81 & 43.69 \\
    MOTIFS~\cite{zellers_neural_2018} &  & \textbf{29.02} & \textbf{38.08} & \textbf{43.64} \\
    \midrule
    Pix2SG B5(Ours) & $\checkmark$ & 19.32 & 23.59 & 25.47\\
    Pix2SG B7(Ours) & $\checkmark$ & 21.51 & 24.81 & 26.66\\
    Pix2SG VIT-B(Ours) & $\checkmark$ & 22.10 & 25.65 & 28.64 \\
    Pix2SG VIT-L(Ours) & $\checkmark$ & \textbf{22.98} & \textbf{26.92} & \textbf{30.05} \\
    \bottomrule
  \end{tabular}
\end{table*}

\begin{table*}
  \centering
  \caption{LF-SGG results on 4D-OR dataset. B5 and B7 represent the corresponding EfficientNet backbones we used in our model.}
  \label{tab:4d_or_results}
  \begin{tabular}{ l | c c | c c c }
  \toprule
    Model & Location-Free & Temporality & Prec. & Rec & F1 \\
    \midrule
    4D-OR baseline~\cite{ozsoy_4d-or_2022}  &  & & 0.68 & 0.87 & 0.75 \\
    LABRAD-OR~\cite{ozsoy2023labrad}  & & $\checkmark$ & 0.87 & 0.90 & 0.88 \\
    \midrule
    Pix2SG B5(Ours)  & $\checkmark$ & & 0.88 & 0.92 & 0.90 \\
    Pix2SG B7(Ours)  & $\checkmark$ & & \textbf{0.89} & \textbf{0.94} & \textbf{0.91} \\
    \bottomrule
  \end{tabular}
\end{table*}

\begin{table*}
  \centering
  \caption{Image retrieval results on Visual Genome. Gallery size refers to the number of images in the image pool from which one image is retrieved.}
  \label{tab:image_ret_results}
  
  \begin{tabular}{ l |  c | c c | c c}
  \toprule
    \multicolumn{2}{c|}{Gallery Size} & \multicolumn{2}{c|}{1000} & \multicolumn{2}{c}{5000} \\
    \midrule
    Model & Location-Free & R@20 & R@100 & R@20 & R@100\\
    \midrule
    MOTIFS~\cite{zellers_neural_2018} &  & 20.8 & 59.2 & 05.2 & 21.3 \\
    VCTree~\cite{tang_learning_2019} & & 19.1 & 55.5 & 05.1 & 20.3 \\
    \midrule
    Pix2SG VIT-L(Ours) & $\checkmark$ & \textbf{38.3} & \textbf{73.9} & \textbf{12.7} & \textbf{39.8} \\
    \bottomrule
  \end{tabular}
\end{table*}

\begin{table*}
  \centering
  \caption{Visual Question Answering results on COCOVQA~\cite{visual_question_answering}.}
  \label{tab:vqa_results}
  \begin{tabular}{ l |  c | c c c | c}
  \toprule
    Model & Location-Free & Open. & Num. & Yes/No & Overall\\
    \midrule
    IMP~\cite{xu_scene_2017}  &  & 26.65 & 32.17 & 66.44 & 42.32 \\
    PSGTR~\cite{yang_panoptic_2022}  & & \textbf{29.03} & \textbf{32.58} & 67.34 & \textbf{43.89} \\
    \midrule
    Pix2SG VIT-L(Ours) & $\checkmark$ & 28.27 & 32.23 & \textbf{67.62} & 43.57 \\
    \bottomrule
  \end{tabular}

\end{table*}

\begin{figure*}[!t]
  \centering
  
  \includegraphics[width=0.85\linewidth]{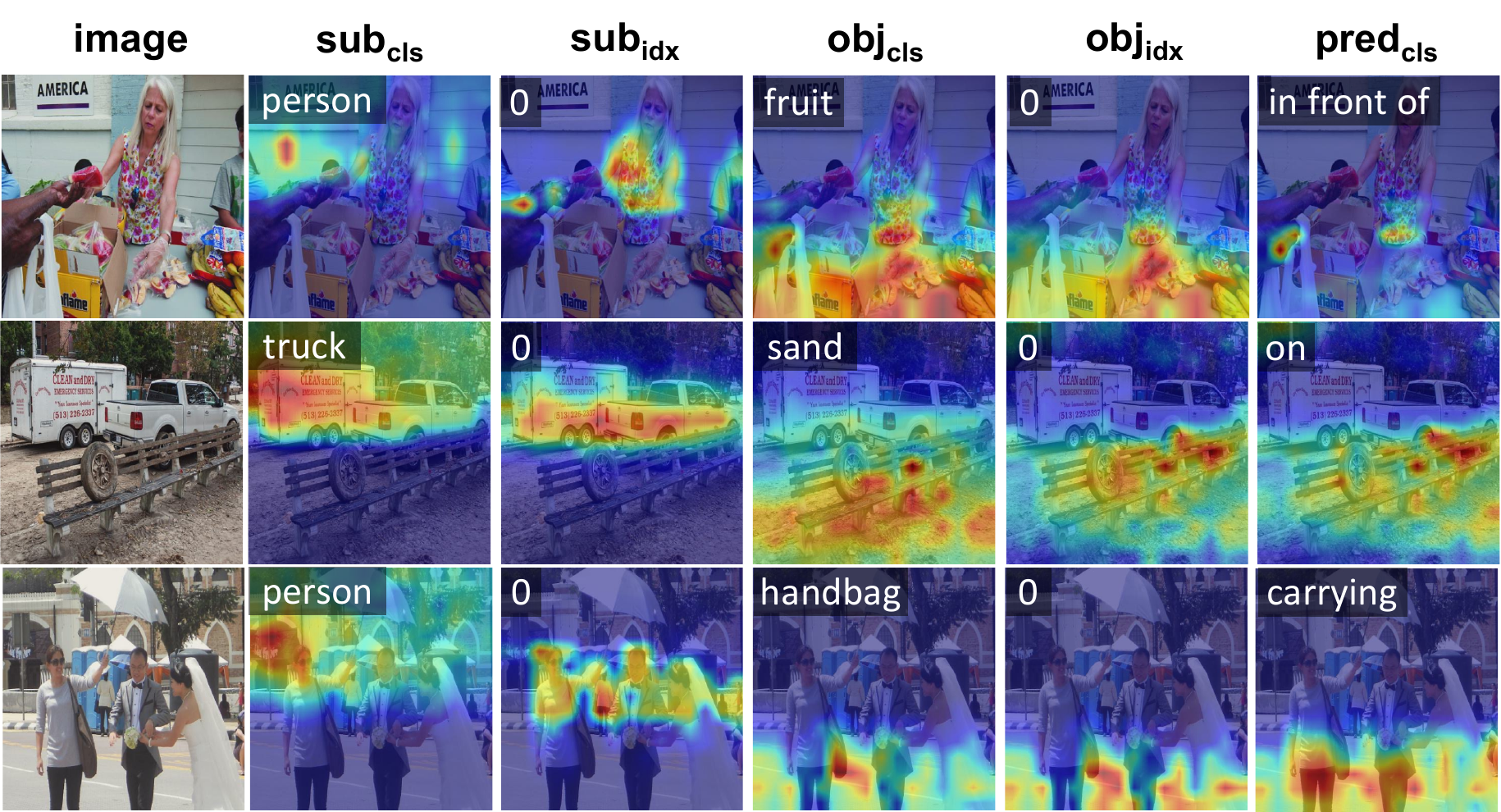}

  \caption{Visualization of the attention maps of Pix2SG on three images. While the subject attention seems to focus on a few entities, the object and predicate attentions tend to focus on the surroundings as well.}
  \label{fig:attention}
\end{figure*}

\begin{table}
  \centering
  \caption{Effect of nucleus sampling with a p-value of 0.95 on VG compared to conventional maximum likelihood selection for LF-SGGen.}
  \label{tab:sampling_results}
  
  \begin{tabular}{l p{0.5cm} p{0.5cm} p{0.75cm}}
  \toprule
    Sampling & R@20 & R@50 & R@100 \\
    \midrule
    Maximum Likelihood & 19.05 & 21.18 & 23.19 \\
    Nucleus \cite{holtzman_curious_2020} & 21.51 & 24.81 & 26.66\\
    \bottomrule
  \end{tabular}

\end{table}

\begin{table}
  \caption{Approximate annotation effort for bounding box labels with median annotation time per bounding box of 42~sec and 8h working days.}
  
  \begin{tabular}{l r r}
  \toprule
    & Visual Genome & 4D-OR \\
    \midrule
    \# of BBox & 1.3M & 189K \\
    hours total & 15,945 & 2,205 \\
    person days & 1,993 & 276 \\
    \bottomrule
  \end{tabular}
  
  \label{tab:annotation_effort}
\end{table}

\begin{table*}
\centering
\caption{Inference speed (FPS) on Visual Genome. For Pix2SG, we show FPS and R@20 for predicting 300, 100 and 20 relations.}

\begin{tabular}{l|c|c|c}
    Method & Location-Free & FPS & R@20\\
    \midrule
    IMP~\cite{xu_scene_2017} &  & 3.00 & 21.66  \\
    MOTIFS~\cite{zellers_neural_2018} & & 2.55 & \textbf{29.02} \\
    Transformer~\cite{tang_unbiased_2020} &  & 3.10 & 28.79 \\
    VCTree~\cite{tang_learning_2019} &  & 1.50 & 27.06 \\
    SS-R-CNN~\cite{teng_structured_2022} &  & \textbf{5.59} & 22.09 \\
    RelTR~\cite{cong_reltr_2022} &  & 4.90 & 25.86 \\
    SGTR~\cite{li_sgtr_2022} &  & 4.98 & 23.62 \\
    \midrule
    Pix2SG 300 & $\checkmark$ & 0.41 & \textbf{21.51} \\
    Pix2SG 100 & $\checkmark$ & 3.90 & 21.31\\
    Pix2SG 20 & $\checkmark$ & \textbf{17.7} & 21.11\\

\end{tabular}

\label{tab:inference_speeds}
\end{table*}

\noindent\textbf{Panoptic Scene Graph Dataset (PSG).} As we introduce a new task and a new metric, we first reevaluate existing methods trained on the task of SGG with segmentation masks, with our LF-SGG evaluation method. While their reliance on masks makes them not directly comparable to our method in the task of LF-SGG, we still provide these results as a rough but valuable guideline. We then evaluate our approach, Pix2SG, which is trained and evaluated without any location labels. We present these results in Tab.~\ref{tab:psg_results}. As we are the first and only method to not require location information at any stage, our method and results serve as the first baseline for the new task of LF-SGG. While using much less annotations, Pix2SG outperforms every method except PSGTR~\cite{yang_panoptic_2022}, validating our location free scene graph generation architecture. In addition to the quantitative results, we also provide qualitative results in Fig.~\ref{fig:qual_psg}, and visualize the attention maps for three quintuple predictions in Fig.~\ref{fig:attention}, illustrating that our model attends to relevant parts of the image for predicting relationships, acting as a sanity check.  When predicting the subject index or object index, the model's attention seems to be very focused on the corresponding object category (e.g. "person 0" in row 3). For the prediction of the predicate, the model attentions seems focused on the contact points between the two entities that were predicted before. These examples highlight the benefits of our autoregressive approach, where the step-by-step prediction of the quintuples allows our model to focus its attention to a single relevant component at a time.

\noindent\textbf{Visual Genome (VG).} Similar to PSG, we also reevaluate existing methods trained on the task of SGG with bounding boxes with our LF-SGG evaluation method. While the results are again not directly comparable, they provide an additional data point. We present these results in Tab.~\ref{tab:vg_results} and Fig.~\ref{fig:qual_vg}. We see that without location supervision, Pix2SG performs comparatively worse on VG than on PSG. We think the discrepancy is mainly caused by lower quality and decreased consistency of the scene graph annotations on Visual Genome, and argue that, as our autoregressive formulation can exploit interrelation dependencies between entities, it thrives where label consistency is maintained.

\noindent\textbf{4D-OR.} As the evaluation proposed in 4D-OR~\cite{ozsoy_4d-or_2022} is applicable to our method, we directly compare our method to the existing results in Tab.~\ref{tab:4d_or_results}. We not only significantly improve upon the existing single frame baseline, from 75\% F1 to 91\% F1, we even outperform the SOTA, which utilizes both visual and temporal information. These results not only support our theory regarding the importance of label consistency, but also validate our approach in a different and unique domain, which signifies the transferability and adaptability of Pix2SG. Importantly, we achieve this without using bounding boxes, depth, or 3D point clouds, which are all used by the existing methods. We provide qualitative results for 4D-OR in Fig.~\ref{fig:qual_4dor}.

\noindent\textbf{Image Retrieval.} In the task of image retrieval, which by design does not use image information, Pix2SG outperforms existing methods by a large margin, even though it uses less labels during training, as can be see in Tab.~\ref{tab:image_ret_results}. This supports our motivation of introducing LF-SGG, as they can be just as useful as location based scene graphs for some downstream tasks.

\noindent\textbf{VQA.} We present the results on the task of zero shot VQA in Tab.~\ref{tab:vqa_results}. We again observe that our method performs very comparable to both location based scene graph generation methods. This again supports the wider use of LF-SGG for downstream tasks not directly requiring accurate pixel location information.

\subsection{Ablation Studies}
\label{sec:ablationstudies}
We perform ablation studies to validate the performance of our evaluation method, as well as to demonstrate the importance of nucleus sampling.

In Fig.~\ref{fig:convergence_20}, we evaluate five different branching factors for our tree search-based matching on the four baseline methods, IMP, VCTree, Motifs, Transformers and Pix2SG on Visual Genome. The results show that our algorithm converges at $B=3$, providing a good balance between speed and matching performance. We, therefore, set $B$ to 3 in all other experiments. 

In Tab.~\ref{tab:inference_speeds}, we present the inference speed of Pix2SG and compare it to other methods on Visual Genome. One interesting aspect of our autoregressive method is that it can be configured to predict any number of relations, offering a speed-accuracy trade off. When predicting only 100 or 20 relations, our model loses little performance, but gains a significant speed boost, which can be valuable in time critical domains and tasks.

In Tab.~\ref{tab:sampling_results}, we show the importance of nucleus sampling by comparing it against always choosing the token with the highest probability. By reducing repetition and increasing variance, it leads to significantly higher recall in all thresholds.

\subsection{Discussion \& Limitations}
The conventional scene graph annotation consists of two sub-tasks, scene graph annotation and bounding box or mask annotation. For the creation of the bounding box labels, the necessary time investment~\cite{su_crowdsourcing_2012} has been reported as 42 seconds for a single bounding box, which breaks down into drawing (25.5~sec), quality verification (9~sec), and coverage verification (7.8~sec).
This allows us to approximate the labeling effort for the dataset used in our experiments which are summarized in Tab.~\ref{tab:annotation_effort}.
The approximated additional workload to create the Bbox labels is significant for both datasets. While it would take a single person 276 days to create the bounding boxes for the 4D-OR dataset, for the Visual Genome dataset, the time required is 1993 days. Even though there exist many methods that can support annotators in improving their efficiency by a factor of 6-9x~\cite{su_crowdsourcing_2012}, this result strongly motivates the design of algorithms that can avoid the use of these costly additional annotations.

Finally, while we reach 90.5\% of the SOTA performance on PSG, our results also indicate that the location cues provide orthogonal information to scene graph labels beneficial for the SGG task. Therefore, this work does not advocate excluding location information when it is accessible. Instead, it presents a novel approach that enables the use of scene graphs even in settings where location information is not available.

\FloatBarrier

\section{Conclusion}
This work introduces the task of location-free scene graph generation (LF-SGG), where we break the dependency of scene graph generation on object location information. Along with this new task, we introduce the first scene graph generation method, Pix2SG, which does not use object location information at any stage of the pipeline. Pix2SG uses autoregressive sequence modeling to directly generate the scene graph from the input image, which reaches competitive results on existing datasets without relying on location information. Furthermore, we provide an efficient heuristic tree search-based algorithm for the complex problem of scene graph matching. This allows the evaluation of LF-SGG without object localization also for future pipelines. Finally, we demonstrate the effectiveness of location free scene graphs on two downstream tasks, image retrieval and visual question answering. We believe that our new task and its objective evaluation can pave the way to more efficient scene interpretation from visual content in the form of scene graphs.

\backmatter

\bmhead{Acknowledgements}
This work has been partially supported by Stryker, the EVUK programme ("Next-generation Al for Integrated Diagnostics”) of the Free State of Bavaria, Bundesministerium für Bildung und Forschung (BMBF) with grant [ZN 01IS17050].

\bmhead{Data Availability}
Datasets used in this work are all publicly available: Visual Genome (VG)~\cite{krishna_visual_2016}, Panoptic Scene Graph Dataset (PSG)~\cite{yang_panoptic_2022}~\url{https://github.com/Jingkang50/OpenPSG}, 4D-OR~\cite{ozsoy_4d-or_2022}~\url{https://github.com/egeozsoy/4D-OR}, COCOVQA~\cite{visual_question_answering}~\url{https://visualqa.org}.

\begin{appendices}
\titleformat{\section}[block]{\large\bfseries}{\appendixname~\thesection:}{1em}{}
\section{Nucleus sampling} 
\begin{figure}
  \centering
  \textbf{Nucleus Sampling Ablation}
   \includegraphics[width=1.0\linewidth]{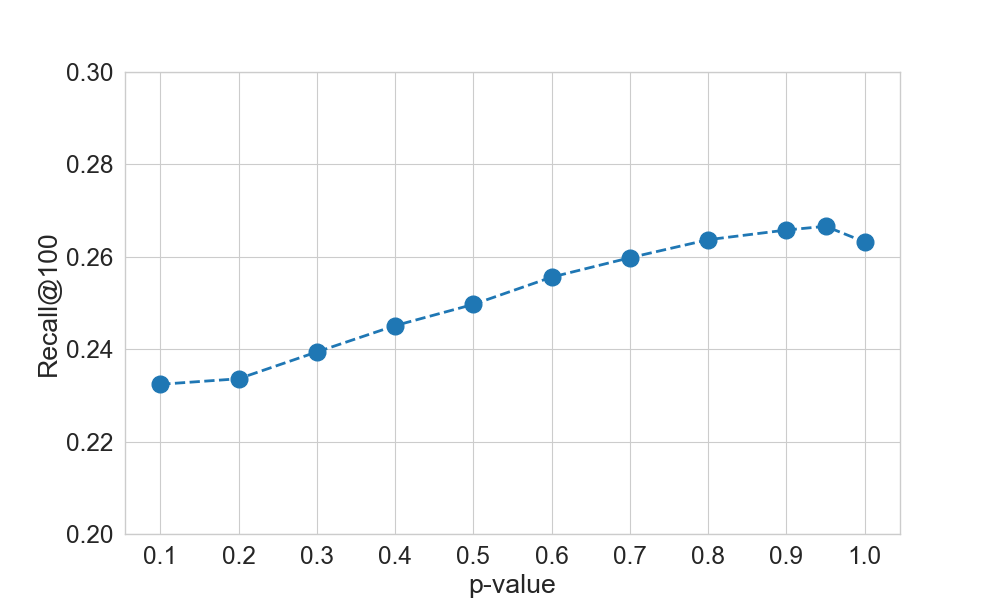}
   \caption{Effect of the p-value for nucleus sampling on R@100 on Visual Genome in the validation set.}
   \label{fig:sg_p_values_100}
\end{figure}

We investigate the effect of different p-values for nucleus sampling~\cite{holtzman_curious_2020} and show the effect on the recall in Fig.~\ref{fig:sg_p_values_100}. The p-value for nucleus sampling determines the classes that are sampled for the next token in the prediction sequence. The classes with the highest probabilities are sampled up to the cumulative probability of p. Higher p-values increase the likelihood of sampling more tokens with lower output probabilities, allowing the model to make more diverse predictions. For lower p-values, only the most probable classes are considered. In our experiments on Visual Genome~\cite{krishna_visual_2016}, we found that the  highest recall is achieved for a p-value of 0.95. 
The improved performance with high p-values indicates that, forcing more diverse predictions is beneficial.

\begin{figure}
  \centering
  
   \includegraphics[width=1.0\linewidth]{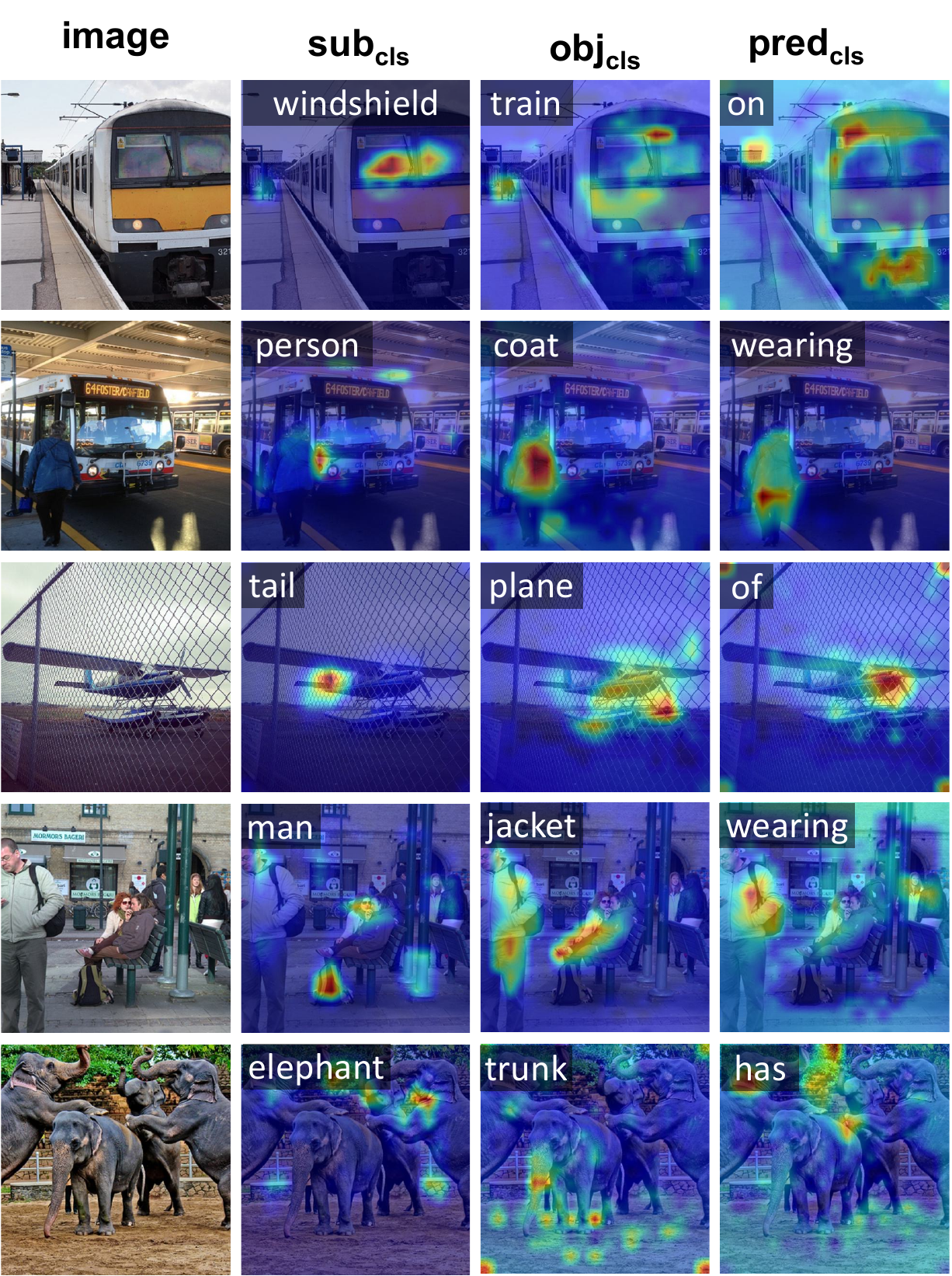}
   \caption{Visualization of our model's attention, illustrating where the model is focusing on while predicting the corresponding subject, object, or predicate.}
   \label{fig:sup_attention}
\end{figure}

\section{Sequence Decoding}
We expand on section II.D of the main paper and provide further details and examples of our encoding and decoding of scene graphs into a sequence of tokens for our Pix2SG autoregressive architecture.
During training, we randomly order the quintuples into a sequence, which is inspired by Pix2Seq~\cite{chen_pix2seq_2022} following a similar procedure for permutationally invariant bounding boxes. E.g. for a ground truth with two sets of quintuples [A,B,C,D,E], [V,W,X,Y,Z], both [A,B,C,D,E,V,W,X,Y,Z] and [V,W,X,Y,Z,A,B,C,D,E] would be valid sequences. This ambiguity is limited through teacher-forcing, where during training, the model is guided by ground truth labels. I.e., the prediction of "X" would be conditioned on the preceding token sequence e.g. [A,B,C,D,E,V,W]. This reduces the uncertainty during training, especially for the latter tokens in the sequence. Empirically, we find that this partially noisy training method works well. For evaluation, we decode the predicted sequence into a scene graph and compare it directly with the ground truth scene graph. Since our evaluation method is entirely based on graphs, it is unaffected by the ordering of the quintuples or the resulting ambiguity, which are specific to our Pix2SG model architecture. This also makes our evaluation algorithm agnostic to the SGG method used.


\section{More Attention Visualizations}
In Fig.~\ref{fig:sup_attention}, we provide more examples of attention maps to show our model's location awareness. Generally, the attention seems to be focused on the location of the corresponding entity.

\begin{figure*}[t]
  \centering
   \includegraphics[width=1.0\linewidth]{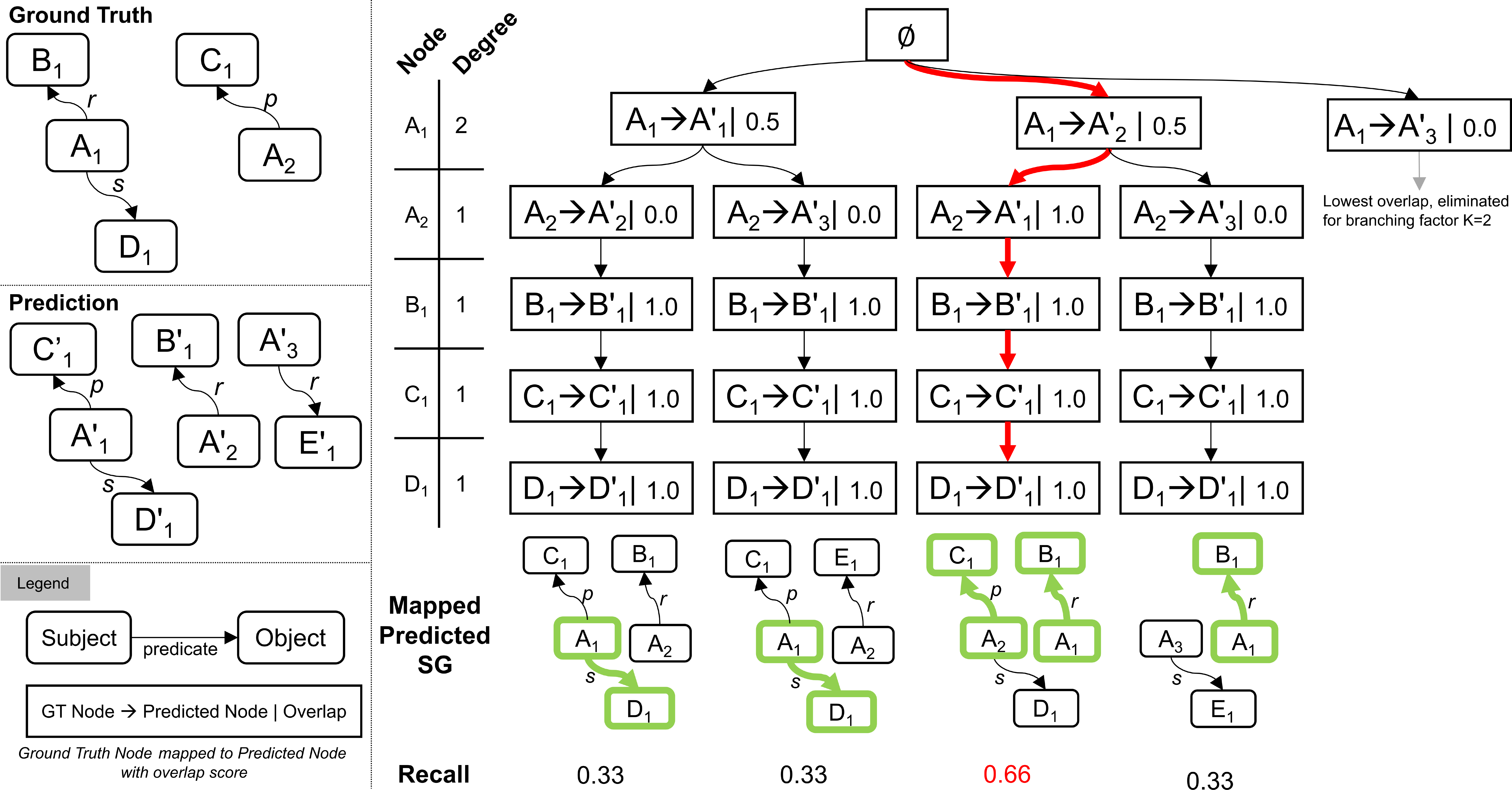}
   \caption{Illustration of our tree-based graph-matching algorithm for a branching factor $K=2$. Given a ground truth scene graph and a predicted scene graph, our algorithm iterates over the ground truth nodes from the highest degree to the lowest in each step and explores $K$ different branches. Afterward, the different pathways are used to calculate the instance match proposals, and the pathway with the highest recall is used to map the predicted scene graph to the ground truth scene graph. For nodes, capital letters denote their class and indices their instance ID, i.e. $"A_1"$ and $"A_2"$  are two instances of class "A".}
   \label{fig:sup_branching}
\end{figure*}

\section{Visualization of the Tree-Based Graph Matching Algorithm}
To provide further insight into our tree-based graph-matching algorithm, we show an exemplary illustration for a branching factor $K=2$ in Fig.~\ref{fig:sup_branching}.

\section{More Qualitative Results on Visual Genome}
In Fig.~\ref{fig:sup_qual_vg}, we provide more examples for the predictions of our model and the corresponding ground truth labels on Visual Genome, and highlight the sparsity and consistency of the annotations. Interestingly, we observe that most predicted entities and predicates seem plausible, even though they are not included in the Visual Genome annotations. In A, the model's predictions describe the visual scene in more detail than the sparse annotations. 
In B, we observe that the model repeats many predictions linked to the entities of \textit{man} and \textit{boy}. In the dataset, similar images can have the entity label \textit{man} or \textit{boy} without a clear distinctive signal. We suspect the model is optimizing for this ambiguity by predicting two separate scene graphs and sets of entities for each of the semantically similar classes.
In C, our model predicts one instance of $<banana,on,bike>$, which can be seen in the scene even though it is not included in the annotations. Interestingly, there is no example of the triplet $<banana,on,bike>$ in Visual Genome, showing our model can generalize to previously unseen semantic connections.

\section{More Qualitative Results on Panoptic Scene Graph Dataset}
\begin{figure*}[t]
  \centering
  
   \includegraphics[width=1.0\linewidth]{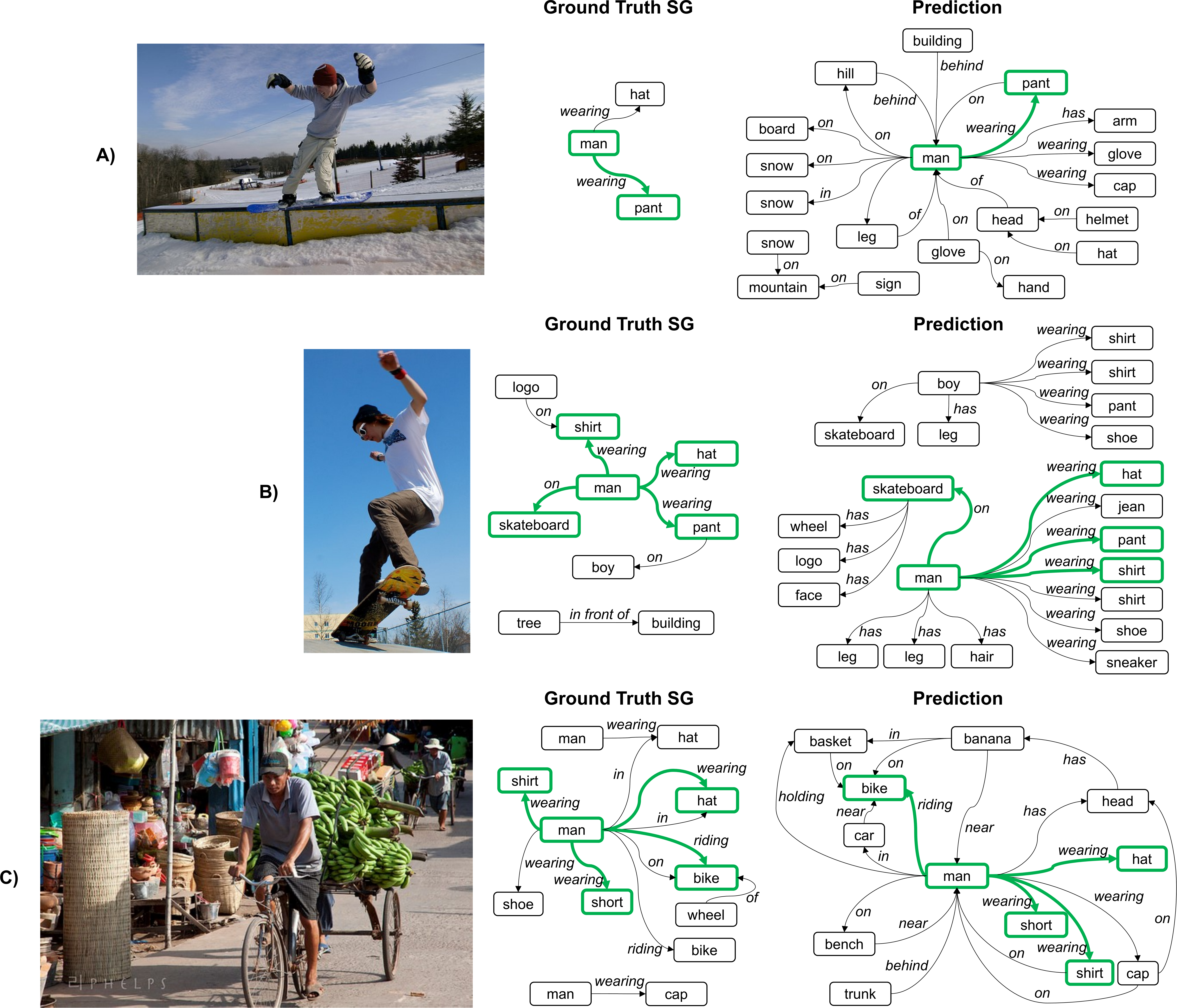}
   \caption{Extended Qualitative Analysis of Pix2SG on Visual Genome. Image, Ground Truth Scene Graph and Predicted Scene Graph are shown. The Predicted Scene Graph is constructed only from the 20 most probable predictions of the model (as in Recall@20). Matching triplets from Ground Truth and Prediction are highlighted in green.}
   \label{fig:sup_qual_vg}
\end{figure*}

\begin{figure*}[t]
  \centering
  
   \includegraphics[width=1.0\linewidth]{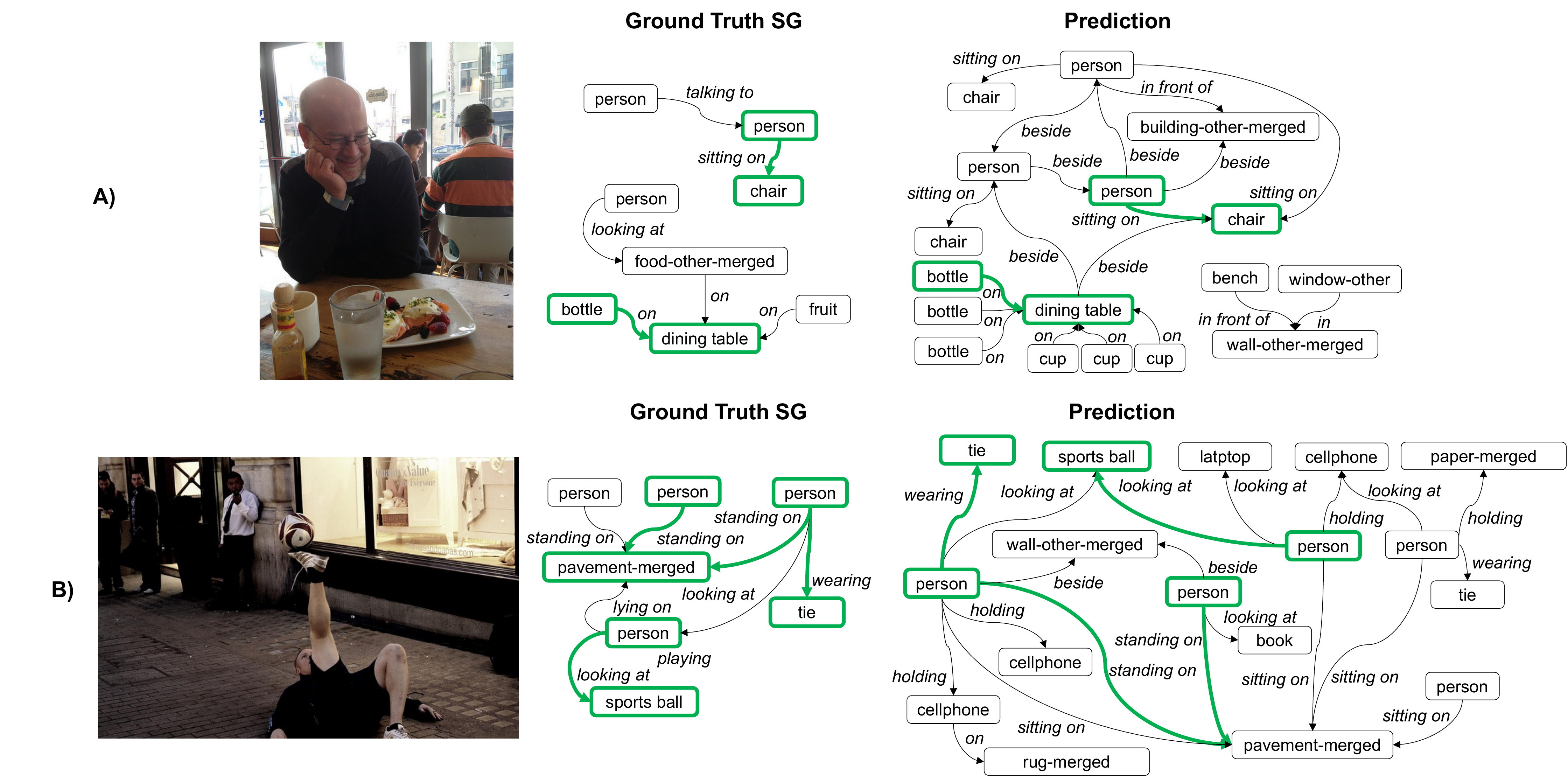}
   \caption{Extended Qualitative Analysis of Pix2SG on Panoptic Scene Graph Dataset. Image, Ground Truth Scene Graph and Predicted Scene Graph are shown. The Predicted Scene Graph is constructed only from the 20 most probable predictions of the model (as in Recall@20). Matching triplets from Ground Truth and Prediction are highlighted in green.}
   \label{fig:sup_qual_psg}
\end{figure*}

In Fig.~\ref{fig:sup_qual_psg}, we provide more examples for the predictions of our model and the corresponding ground truth labels on the panoptic scene graph dataset. Compared to visual genome, the annotations are more comprehensive, yet there are still many predictions of our model that seem plausible but are not part of the ground truth annotations. We do not observe the same behaviour as in Fig.~\ref{fig:sup_qual_vg} B) where the model is unsure of the correct classification ("boy" vs. "man") and therefore predicts multiple scene graphs, as these classes are combined into the class "person". The model can, however, still struggle to predict the correct number of entities, as seen in Fig.~\ref{fig:sup_qual_psg} A) ("bottle" and "cup").

\section{Qualitative Image Generation Results}
\begin{figure*}[t]
  \centering
  
   \includegraphics[width=1.0\linewidth]{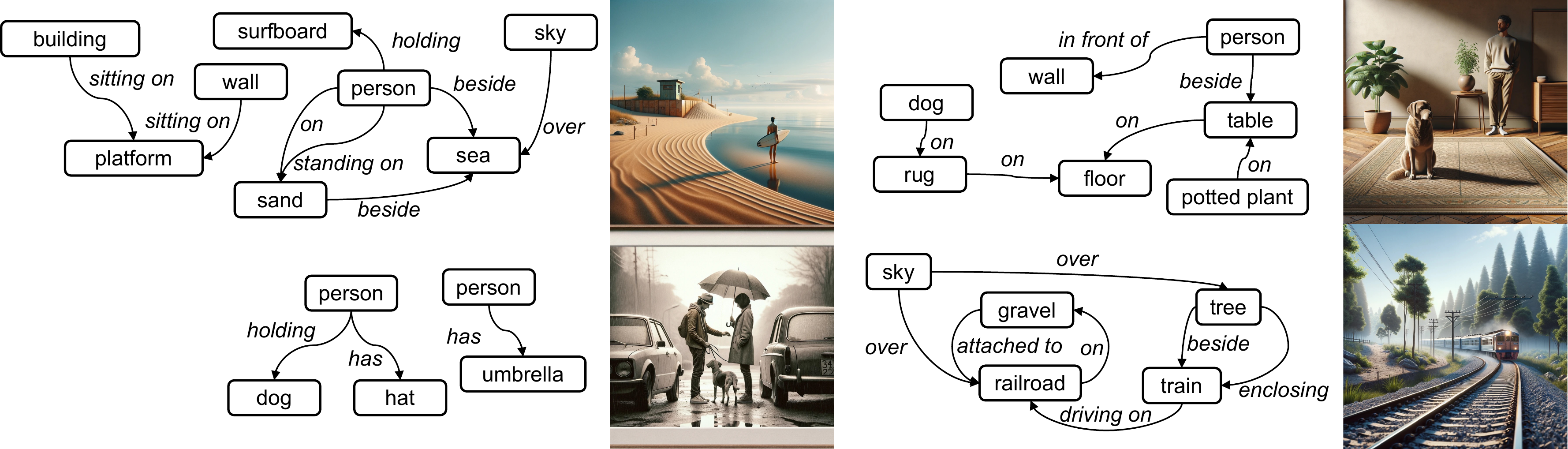}
   \caption{Qualitative examples of image generation from location-free scene graphs.}
   \label{fig:sup_image_gen}
\end{figure*}

In Fig.~\ref{fig:sup_image_gen}, we include qualitative examples for a third downstream task, image generation. We generate an image from an LF-SG, by using GPT-4 Vision, a model not trained for this task. We prompt it with the following prompt "A realistic real-world photo that matches the following scene graph. The photo should not have any details not mentioned here: \textless LF-SG\textgreater", where \textless LF-SG\textgreater~is a location-free scene graph represented as a list of quintuplets. These first results indicate that location-free scene graphs could also be a potent representation for image generation tasks.

\end{appendices}

\FloatBarrier
\balance
\bibliographystyle{apalike.bst} 
\bibliography{sn-bibliography}


\end{document}